\pdfoutput=1

\documentclass{article}

\usepackage{arxiv}
\usepackage{natbib}
\usepackage{natbib}
\bibpunct[, ]{(}{)}{,}{a}{}{,}%

\usepackage{hyperref}

\usepackage{times}
\usepackage{latexsym}
\usepackage{booktabs}
\usepackage{multirow}
\usepackage{amsmath}
\usepackage{float}
\usepackage{subcaption}
\usepackage{comment}
\usepackage{array}
\usepackage[table, svgnames, dvipsnames]{xcolor}
\usepackage[T1]{fontenc}

\usepackage[utf8]{inputenc}

\usepackage{microtype}

\usepackage{inconsolata}

\usepackage{graphicx}

\title{Bridging the LLM Accessibility Divide? Performance, Fairness, and Cost of Closed versus Open LLMs for Automated Essay Scoring}

\author{ \textbf{Kezia Oketch,\textsuperscript{1}} \textbf{John P. 
Lalor,\textsuperscript{1}} \textbf{Yi Yang,\textsuperscript{2}}
  \textbf{Ahmed Abbasi\textsuperscript{1}}
\\
\\
  \textsuperscript{1}Department of IT, Analytics, and Operations,
  University of Notre Dame \\
 \textsuperscript{2}Department of Information Systems, Business Statistics and Operations Management,\\ Hong Kong University of Science and Technology
\\
  \small{ \texttt{\{koketch, john.lalor, aabbasi\}@nd.edu}, \texttt{imyiyang@ust.hk} 
  }
}

\begin{document}
\maketitle
\begin{abstract}
Closed large language models (LLMs) such as GPT-4 have set state-of-the-art results across a number of NLP tasks and have become central to NLP and machine
learning (ML)-driven solutions.
Closed LLMs' performance and wide adoption has sparked considerable debate about
their accessibility in terms of availability, cost, and transparency. 
In this study, we
perform a rigorous comparative analysis of nine leading LLMs, spanning
closed, open, and open-source LLM ecosystems, across text assessment and
generation tasks related to automated essay scoring. 
Our findings reveal that for few-shot learning-based assessment of human
generated essays, open LLMs such as Llama 3 and Qwen2.5 perform comparably to GPT-4 in terms of predictive performance, with no significant
differences in disparate impact scores when considering age- or race-related fairness.
Moreover, Llama 3 offers a substantial cost advantage, being up to 37 times more
cost-efficient than GPT-4. For generative tasks, we find that essays generated
by top open LLMs are comparable to closed LLMs in terms of their semantic
composition/embeddings and ML assessed scores. Our findings challenge the
dominance of closed LLMs and highlight the democratizing potential of open LLMs,
suggesting they can effectively bridge accessibility divides while
maintaining competitive performance and fairness.
\end{abstract}

\section{Introduction}
The rapid development of machine learning (ML) technologies, particularly large
language models (LLMs), has led to major advancements in natural language
processing \citep[NLP,][]{abbasi2023data}. While much of this advancement
happened under the umbrella of the common task framework which espouses
transparency and openness \citep{abbasi2023data}, in recent years, closed LLMs
such as GPT-3 and GPT-4 have set new performance standards in tasks ranging from text
generation to question answering, demonstrating unprecedented capabilities in
zero-shot and few-shot learning scenarios
\citep{brown2020languagemodelsfewshotlearners,openai2023gpt}. Given the strong
performance of closed LLMs such as GPT-4, many studies within the LLM-as-a-judge
paradigm rely on their scores as ground truth benchmarks for evaluating both
open and closed LLMs \citep{chiang2023can}, further entrenching the dominance of
SOTA closed LLMs  \citep{vergho2024comparing}. 
Along with closed LLMs, there are also LLMs where the pre-trained models (i.e., training weights) and inference code are
publicly available (``open LLMs'') such as Llama \citep{touvron2023llama,dubey2024llama} as well as LLMs  where the full training data and training code are also available (``open-source LLMs'') such as OLMo \citep{groeneveld2024olmo}.
Open and open-source LLMs provide varying levels of transparency for developers and researchers~\citep{liu2023llm360}.

Access to model weights, training data, and inference
code enables several benefits for the user-developer-researcher community,
including lower costs per input/output token through third-party API services,
support for local/offline pre-training and fine-tuning, and deeper analysis of
model biases and debiasing strategies. However, the dominance of closed LLMs
raises a number of concerns, including accessibility and fairness
\citep{strubell2020energy,bender2021bender,irugalbandara2024scaling}. The
accessibility divide in this context can be understood in three dimensions:
uneven availability due to geographic and economic barriers, prohibitive costs
that limit adoption, and a lack of transparency that hinders research and
innovation. 

In the LLM space, corporate-driven commodification---through
monopolized APIs and exclusive licensing---is exacerbating the accessibility
divide~\citep{luitse2021great,abbasi2024pathways}. These challenges are both
technical and ethical, impacting who can access and benefit from the
opportunities afforded by SOTA LLMs---those affected include researchers and
practitioners residing in less affluent regions and/or complex socio-political
environments. Open and open-source LLMs such as Llama 3, Qwen2.5, and OLMo 2 provide
greater transparency and customization potential
\citep{touvron2023llama,dubey2024llama,Bai2023QwenTR,groeneveld2024olmo}. As
these models improve in general benchmarking tasks, there is a need to
systematically compare open and open-source LLMs with their closed SOTA
counterparts on different assessment/scoring and generation tasks across various
dimensions including performance and fairness. We aim to address this gap by
conducting a comprehensive comparative analysis of nine LLMs---encompassing
closed, open, and open-source LLMs---across multiple text generation and
evaluation tasks. The Research Questions (RQs) guiding this study are:
\textbf{RQ1}: How do different generations of open, open-source and closed LLMs
compare in their assessment capabilities?
\textbf{RQ2}: When performing assessments/scoring, to what extent do closed and
open LLMs exhibit biases?
\textbf{RQ3}: How comparable are open and open-source LLMs to their closed
counterparts in terms of text generation capabilities?

To answer these questions, we use automated essay scoring (AES) as our focal
context. AES is well-suited for our research questions; it has been studied
extensively by the NLP community \citep{ke2019automated},
entails prompt-guided text generation, has readily available large-scale human
testbeds with demographic information, and includes well-defined
evaluation rubrics.  

Our contributions are three-fold: (1) we provide empirical evidence of the
trade-offs between accuracy, cost, and fairness for LLMs when performing
assessment/scoring tasks; (2) we statistically and visually demonstrate the text
generation capabilities of leading open, open-source, and closed LLMs; (3) we
highlight the growing viability of open and open-source LLMs as cost-effective
alternatives to closed LLMs. To the best of our knowledge, this is the first
study to compare the three LLM ecosystems---closed, open, and open-source---across
multiple assessment and text generation tasks.\footnote{Code and data are available at \url{https://github.com/nd-hal/llm-accessibility-divide}.}

\section{Related Work}

\subsection{LLMs and Accessibility}
Accessibility concerns can manifest in many ways, including
the ability to serve those with physical impairments or cognitive impediments.
Here, following prior work, we focus on accessibility as it relates to
availability, cost, and transparency \citep{luitse2021great,abbasi2024pathways}.
Until recently, much of the progress in NLP representation learning and language modeling over the past 20 years occurred under the common task framework and transpired
via publicly available, open and open-source LLMs, methods, algorithms,
architectures, and systems \citep{abbasi2024pathways, abbasi2023data}. New
proprietary LLMs such as GPT-4 are less available in lower- and
middle-income countries due to inadequate internet penetration,
underdeveloped infrastructure, and/or strict censorship policies \citep{wang2023chatgpt}. 

Moreover,
cost-efficiency is a critical factor influencing the adoption of LLMs for
various NLP tasks. \citet{strubell2020energy} examined the environmental and
financial costs associated with training LLMs like GPT-3. Their findings suggest
that the high costs are not only a barrier to accessibility but also raise
concerns about the sustainability of such models. Furthermore, proprietary
models like GPT-4, despite their strong performance, limit researchers' ability
to scrutinize and mitigate biases due to their closed nature
\citep{raji2020closing,bommasani2021opportunities, liao2023ai}. In contrast,
open and open-source LLMs, with their publicly available model weights and
training data/code, offer greater traceability and scrutiny
\citep{eiras2024risks}. 

\subsection{The Performance of Open, Open-source, and Closed LLMs}
The strong performance of closed LLMs such as GPT-3.5
and GPT-4 has led to their adoption as stand-in proxies for human assessors for
ground-truth evaluation \citep{chiang2023can}. Such models have been used as 
judges in various studies related to the evaluation of open-ended tasks
\citep{L-eval_an-etal-2024-l}. For instance, \citet{zheng2023judging} found
models such as GPT-4 can yield agreement rates of up to 80\% with human experts.
However, the growing capabilities of open and open-source LLMs warrant a
systematic comparison.

Prior work highlights that while closed LLMs often lead in terms of raw performance, open and
open-source LLMs offer substantial cost advantages, making them more accessible
to a wider range of users~\citep{irugalbandara2024scaling,kukreja2024literature}. Recently, \citet{wolfe2024laboratory} examined the
impact of fine-tuning smaller open LLMs versus employing
few-shot learning for larger closed LLMs. Their results were mixed; for certain text classification
problems, fine-tuning two open LLMs, Llama-2-7b and Mistral-7b, led to 
performance comparable to few-shot learning with GPT-4. For some other
tasks, the fine-tuned closed LLMs attained markedly better classification
performance. We build on this emergent literature by comparing open, open-source, and closed LLMs in terms of their generation, few-shot
assessment/scoring, and fairness capabilities.

\subsection{Automated Essay Scoring and LLMs}
Automated Essay Scoring (AES) entails rule-based or ML model-based assessment of
human-generated essays in response to different genres of prompts.
Essays
are scored against a defined evaluation rubric focusing on overall essay
quality and/or aspect-oriented quality \citep{ke2019automated,
attali2006automated}. NLP models for AES have evolved from
feature-based ML to RNN/CNN-based deep learning to the use of fine-tuned or
few-shot-learned language models
\citep{ke2019automated,taghipour2016neural,bevilacqua2023automated}.

While AES models have improved, concerns about fairness and bias in AES have persisted.
\citet{ke2019automated} highlighted that AES models could inadvertently
reinforce biases present in training data, including those related to
socioeconomic background or language proficiency. \citet{schaller2024fairness}
explored strategies for mitigating such biases to ensure that AES systems
produce fair and equitable scores. \citet{bevilacqua2023automated} examined the
behavior of ML assessment models scoring human- versus LLM-generated essays and
found that assessors such as BERT and RoBERTa may exhibit a familiarity bias
when scoring LLM-generated essays. As noted in the introduction, we use AES as
our focal context to compare open and closed LLMs because of the familiarity of
the problem to the NLP community, 
availability of large-human-generated text corpora,
presence of different genres of text with clear prompts, and well-defined
instructions and evaluation rubrics.

\section{Data, Models, and Experiments}
\begin{figure*}[tbh]
        \centering
    \includegraphics[width=\textwidth]{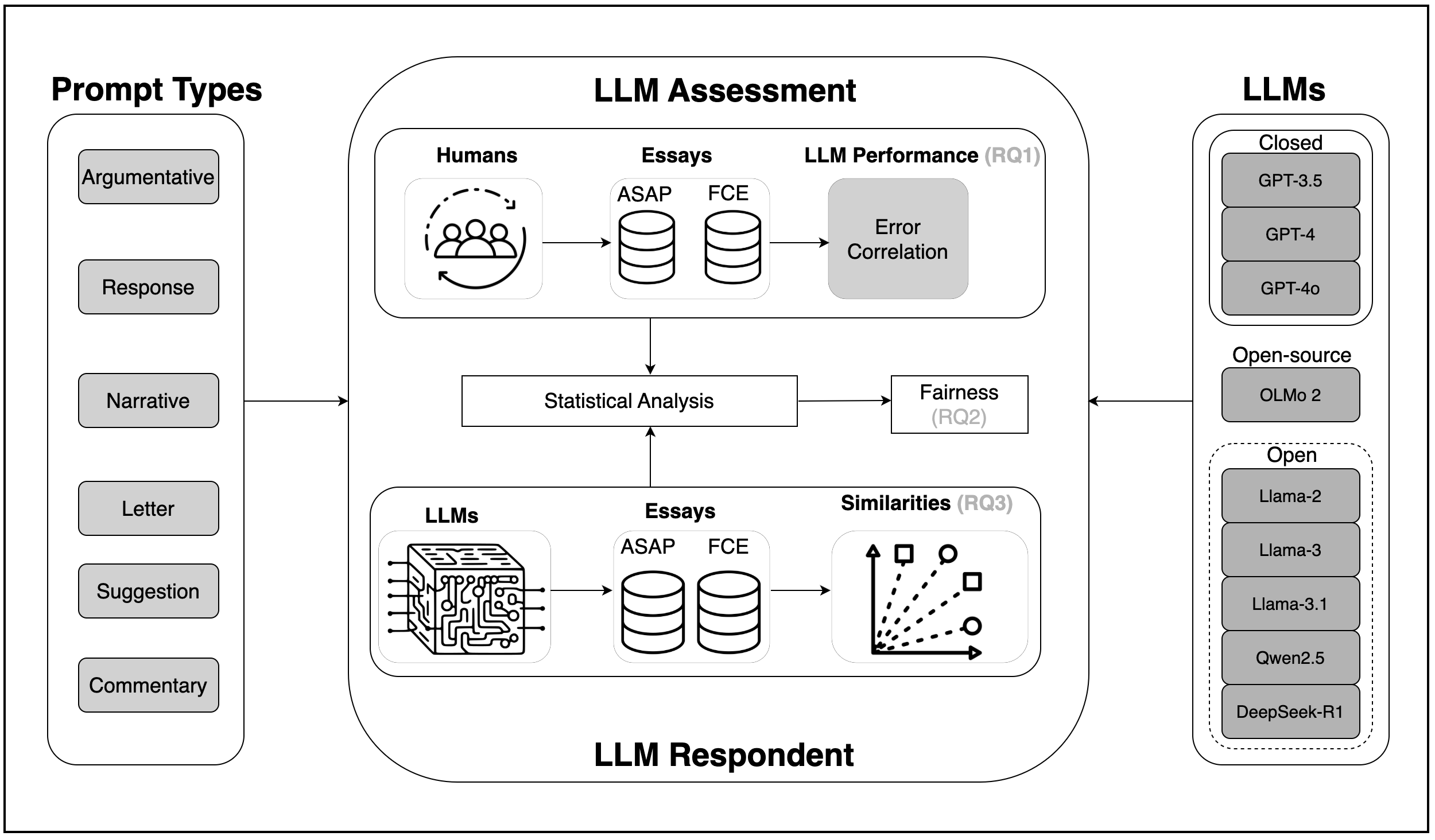}
        \caption{Analysis Framework}
        \label{fig:framework}
\end{figure*}

To answer our three research questions, we developed a robust analysis framework 
(Figure~\ref{fig:framework}). In the remainder of the
section, we describe the data, models, and experiments in detail. 

\subsection{Human Text Data and Prompts}
We use two human-generated essay datasets the Automated Student Assessment
Prize \citep[ASAP,][]{mathias2018asap++} and the Cambridge Learner Corpus-First
Certificate in English exam \citep[FCE,][]{yannakoudakis2011new}.
The ASAP dataset is
widely used as a benchmark dataset in the AES field
\citep{taghipour2016neural,jin2018tdnn}, and
consists of 12,979 essays across 8 prompts (Table \ref{tab:asap_aeg}).
For all essays, we use the overall quality score. 
FCE is a large collection of texts produced by English language learners
from around the world.
Like ASAP, FCE is a widely recognized resource
in NLP that has been used in previous benchmarking
studies \citep{ramesh2022automated,ke2019automated}. 
FCE assesses English at an
upper-intermediate level. 
Test-takers were prompted to complete two writing tasks: a letter, a report, an article, a
composition, or a short story.
For each test-taker a composite score was given across the two tasks. 
FCE is comprised of 2,466 essays spanning 5 genres.

\begin{table}[tb]
\centering
\small 
\begin{tabular}{ccccc}
\toprule
\textbf{Data} & \textbf{Essay Type} & \textbf{N} & \textbf{Avg. Length} & \textbf{Score} \\ \midrule
\textbf{ASAP}&&&&  \\
1 & A & 1784 & 350 & 1 - 6 \\ 
  2 & A & 1800 & 350 & 1 - 6 \\ 
  3 & R & 1726 & 150 & 0 - 3 \\ 
  4 & R & 1772 & 150 & 0 - 3 \\ 
  5 & R & 1805 & 150 & 0 - 4 \\ 
  6 & R & 1800 & 150 & 0 - 4 \\ 
  7 & N & 1569 & 300 & 0 - 30 \\ 
  8 & N & 723 & 650 & 0 - 60 \\ \midrule 
\textbf{FCE} &&&&\\
1 & L &
1237 & 200-400 & 0 - 40 \\ 
  2 & A,C,N,S & 362 & 200-400 & 0 - 40 \\ 
  3 & A,C,L,N & 340 & 200-400 & 0 - 5 \\ 
  4 & A,C,L,N & 498 & 200-400 & 0 - 5 \\ 
  5a & A,C,L,S & 15 & 200-400 & 0 - 5 \\ 
  5b & A,C,L & 14 & 200-400 & 0 - 5 \\ 
 \bottomrule
\end{tabular}
\caption{Description of the data used in this study.  \textit{Avg. Length} gives
the average essay length in number of words. \textit{Score} lists the scoring
range of the various essays. Essay types: argumentative (A), commentary (C),
letter (L), suggestion (S), narrative (N), response (R).}
\label{tab:asap_aeg}
\end{table}

As depicted in Figure \ref{fig:framework}, we use these testbeds, including the
evaluation rubrics, directly as the input data for zero/few-shot-based LLM
assessment (RQ1 and RQ2). We also use the six prompt types and associated
instructions to generate essays with LLM respondents (RQ3).

\subsection{Using LLMs for Assessment}
Following prior work on zero and few-shot in-context learning
\citep{chiang2023can,chen2023exploringuselargelanguage,duan2024exploring}, and
based on our first research question (RQ1), we evaluate the quality of text
written by humans using LLMs for assessment/scoring. We present the LLM with the
task instruction, description of the rating task, rating criteria, the sample to
be rated, and a sentence that prompts the LLM to give the rating. The
instructions, description, and rating critera are presented exactly as they
appear in our corpora. The rating sentence at the end of the prompt
asks the LLM to rate the overall sample quality using a specified scale based
on the original scoring range (Table \ref{tab:asap_aeg}). We tested two settings: 
zero-shot, where no example essays were provided, and few-shot, where in
addition to the rubric and task instructions, three randomly selected human
essays were provided along with their human expert ratings.\footnote{We did not include OLMo 2 in the few-shot assessment task, as its smaller context window (4k) meant a large number of few-shot cases would have been excluded.} We intentionally
selected one random sample per tertile from the human scoring range. LLM scores
were normalized to a 0-1 range. 

Consistent with RQ1, we compare the performance of LLMs for assessing
human-generated text. Following prior research
\citep{bevilacqua2023automated,ramesh2022automated, ke2019automated}, two
categories of metrics were utilized. The first category comprised of two error
metrics: mean squared error (MSE) and mean absolute error (MAE). The second
category comprised of agreement and correlational metrics, specifically
Quadratic Weighted Kappa (QWK), Pearson correlation coefficient (PCC), and
Spearman's rank correlation (SRC). 

\subsection{LLMs as Respondents: Generating Textual Data}
We followed prior work when designing our prompts for LLM essay generation \citep{bevilacqua2023automated,zheng2023can}. Specifically, we used the 
superset of prompts seen by human respondents across the ASAP and FCE.
This resulted
in nearly 150 prompts associated with 68 prompt IDs. To better align
with a human text generation process, we used a zero-shot setting where the LLMs
were provided the exact same instructions as humans, and did not see example
essays as part of the prompts. For the GPT
models, we provided essay prompts via the OpenAI API. For the Llama models, we
used the Replicate API for Llama 2 and Llama 3, and the Llama API for Llama 3.1.
For Qwen2.5 and DeepSeek-R1, we used DeepInfra API. OLMo 2 was run
locally. Each prompt was provided to the LLM 10 times resulting in 1,537 total
essays for each model.\footnote{GPT-4 and GPT-4o failed to respond to two/one of the 68 prompts
resulting in 1,486 and 1,527 essays, respectively.} 
The LLM-generated essays are depicted in the bottom part of Figure
\ref{fig:framework} under ``LLM Respondent'' and inform our third research
question (RQ3). 

\subsection{Statistical Analysis}
For both RQ2 and RQ3, as noted in Figure \ref{fig:framework}, we used
statistical models to allow us to parsimoniously examine the fairness and
generation capabilities of open and closed LLMs while controlling for the types
of prompts, specific prompt IDs, and assessment models.
\subsubsection{Statistical Analysis for Fairness}
\label{ssec:statFairness}
For RQ2, we wanted to examine the fairness of the LLM assessors while
controlling for prompt types/IDs, and the various assessment models. To achieve
this, we ran a three-way ANOVA (split-plot design). We focused solely on
human-generated essays appearing in the FCE corpus due to the availability of
demographic information about the human authors. Following prior work, 
we define bias as representational harm from model
error attributed to protect attributes such as demographics~\citep{lalor2024should}. 
We used the available
demographics in FCE, age ($a$) and race ($r$), as independent
variables in separate ANOVA models. We also include prompt type ($p$) as an
independent variable, as well as the assessment LLM employed ($s$); we also control
for the specific prompt ID ($d$). The dependent variable ($\Delta_\text{R}$) is the
difference between the actual ground truth quality score for the essay ($z$), and
the LLM score ($\hat{z}$). Hence, the statistical fairness ANOVA model is as
follows:    

\begin{align*}
    \Delta_{\text{R}_{ijk}} &= \frac{p_i}{d} + p_i + a_j + s_k + (pa)_{ij} + (ps)_{ik} + 
    (as)_{jk} + (pas)_{ijk} + \epsilon_{ijk} & \text{\textit{age}} \\ 
    \Delta_{\text{R}_{ijk}} &= \frac{p_i}{d} + p_i + r_j + s_k + (pr)_{ij} + (ps)_{ik} + 
    (rs)_{jk} + (prs)_{ijk} + \epsilon_{ijk} & \text{\textit{race}} \\ 
\label{eqn:fair_anova}
\end{align*}

Where $\Delta_\text{R} = z - \hat{z}$, $a$ is binarized into Young (25 and below) and Old (26 and above),
$r$ is
binarized by race (Asian and Non-Asian), $i$,$j$,$k$ refer to
the factor category levels for $p$,$a$,$s$, respectively, and $\epsilon$ is the
random error term.   

\subsubsection{Statistical Analysis for Response Generation}
\label{ssec:statGen}

For RQ3, we wanted to examine the response generation commonalities and
differences of various open and closed LLMs relative to one another and humans.
Similar to the fairness statistical model, here, we controlled for prompt
types/IDs, and the various assessment models. To achieve this, we ran another
three-way ANOVA (split-plot design) setup. We
used the full set of essays generated by humans (ASAP and FCE) and the six LLMs
(across all ASAP/FCE prompts). The dependent variable is the assessment LLM
score ($\hat{z}$). Instead of demographics, we use $t$ to indicate the respondent
type with seven possible values: one of the six LLMs or human. Once again, we
include prompt type ($p$) as an independent variable, as well as the assessment
LLM employed ($s$), and control for the prompt ID ($d$). Hence, the statistical
response generation model is as follows:    

\begin{equation*}
\label{eqn:resp_anova}
\begin{aligned}
    \hat{z} = \frac{p_i}{d} + p_i + t_j + s_k + (pt)_{ij} + (ps)_{ik} +
    (ts)_{jk} + (pts)_{ijk} + \epsilon_{ijk}
\end{aligned}
\end{equation*}

Where $i$,$j$,$k$ refer to the factor category levels for $p$,$t$,$s$,
respectively, and $\epsilon$ is the random error term.

\section{Results}
\subsection{Performance of LLMs for Assessment} 
Related to RQ1, we evaluated the assessment/scoring performance of LLMs when
evaluating human-generated text with expert ground-truth labels. We present our
benchmarking results in Table \ref{tab:performance_with_source_type}. Each of
the nine LLMs was presented with both human-generated and LLM-generated text. As
noted, the dependent variable was normalized to a continuous scale ranging from
0 to 1. We applied two error metrics, MSE and MAE, along with three agreement
and correlation measures, QWK, PCC, and SRC
\citep{bevilacqua2023automated,ramesh2022automated,ke2019automated}. For closed
LLMs, GPT-4 demonstrated the best performance in both zero-shot and few-shot
settings, followed by GPT-4o and GPT-3.5, respectively.

For open LLMs, Qwen2.5-72B achieved the highest overall performance, followed by
Llama 3, DeepSeek-R1, Llama 3.1, and Llama 2, in both zero-shot and few-shot
conditions. Notably, the performance gap between zero-shot and few-shot settings
is narrower for open LLMs compared to closed LLMs, suggesting that open models
may be more stable across inference settings or benefit less from few-shot
learning.
In particular, Qwen2.5 (FS) and Llama 3 (FS) are highly competitive with GPT-4 (FS). Qwen2.5 outperformed GPT-4 on MSE (0.051 vs. 0.045) and MAE (0.159 vs. 0.163), Llama 3 outperformed GPT-4 on QWK (0.970 vs. 0.964) while achieving
comparable results on PCC and SRC. This highlights that certain open
models are closing the performance gap with state-of-the-art closed models in
structured evaluation tasks.

For the open-source LLM, OLMo 2-13B was evaluated in a zero-shot setting only.
While its performance lags behind closed and open models—particularly in QWK
(0.767)—it remains competitive in correlation metrics (PCC: 0.357, SRC: 0.360),
outperforming some open models in their zero-shot configurations. This suggests
that, although open-source models may currently trail behind leading LLMs, they
offer a viable alternative for users prioritizing transparency, cost-efficiency,
and local deployment.

\begin{table}[tb]
\centering
\small 
\begin{tabular}{lccccc}
\toprule
\textbf{Model}  & \multicolumn{2}{c}{\textbf{Errors $\downarrow$}} & \multicolumn{3}{c}{\textbf{Correlations $\uparrow$}} \\
   & \textbf{MSE} & \textbf{MAE} & \textbf{QWK} & \textbf{PCC} & \textbf{SRC}
 \\
\midrule
\multicolumn{6}{c}{\textbf{Closed LLMs}} \\
\rowcolor{Gainsboro!60}
GPT-3.5  & .199 & .362 & .573 & .251 & .238 \\
 & .070 & .198 & .960 & .443 & .419 \\
\rowcolor{Gainsboro!60}
GPT-4  & .053 & .163 & .951 & \textbf{.595} & \textbf{.596} \\
 & .052 & \textbf{.163} & .964 & .553 & .534 \\
\rowcolor{Gainsboro!60}
GPT-4o  & .183 & .359 & .586 & .309 & .336 \\
 & .056 & .179 & .960 & .580 & .570 \\
\midrule
\multicolumn{6}{c}{\textbf{Open LLMs}} \\
\rowcolor{Gainsboro!60}
Llama 2 70B & .157 & .328 & .858 & .049 & .024 \\
 & .080 & .226 & .961 & .203 & .238 \\
\rowcolor{Gainsboro!60}
Llama 3 70B  & .052 & .169 & .924 & .504 & .511 \\
  & .051 & .168 & \textbf{.970} & .548 & .532 \\
\rowcolor{Gainsboro!60}
\scriptsize{Llama 3.1 405B}  & .059 & .164 & .925 & .525 & .517 \\
 & .065 & .188 & .961 & .441 & .421 \\
\rowcolor{Gainsboro!60}
DeepSeek-R1 671B  & .054 & .171 & .899 & .436 & .430 \\
 & .069 & .193 & .943 & .349 & .346 \\
\rowcolor{Gainsboro!60}
Qwen2.5-72B  & .051 & .166 & .919 & .492 & .493 \\
  & \textbf{.045} & \textbf{.159} & .955 & \textbf{.560} &
 \textbf{.546} \\
\midrule
\multicolumn{6}{c}{\textbf{Open-Source LLM}} \\
\rowcolor{Gainsboro!60}
OLMo 2-13B  & .073 & .209 & .767 & .357 & .360 \\
\bottomrule
\end{tabular}
\caption{Performance metrics for benchmark models across the two datasets under zero-shot (shaded rows) and few-shot (unshaded rows) settings.}
\label{tab:performance_with_source_type}
\end{table}

\begin{figure}
    \centering
    \begin{subfigure}[t]{\columnwidth}  
        \centering
        \includegraphics[width=0.8\columnwidth]{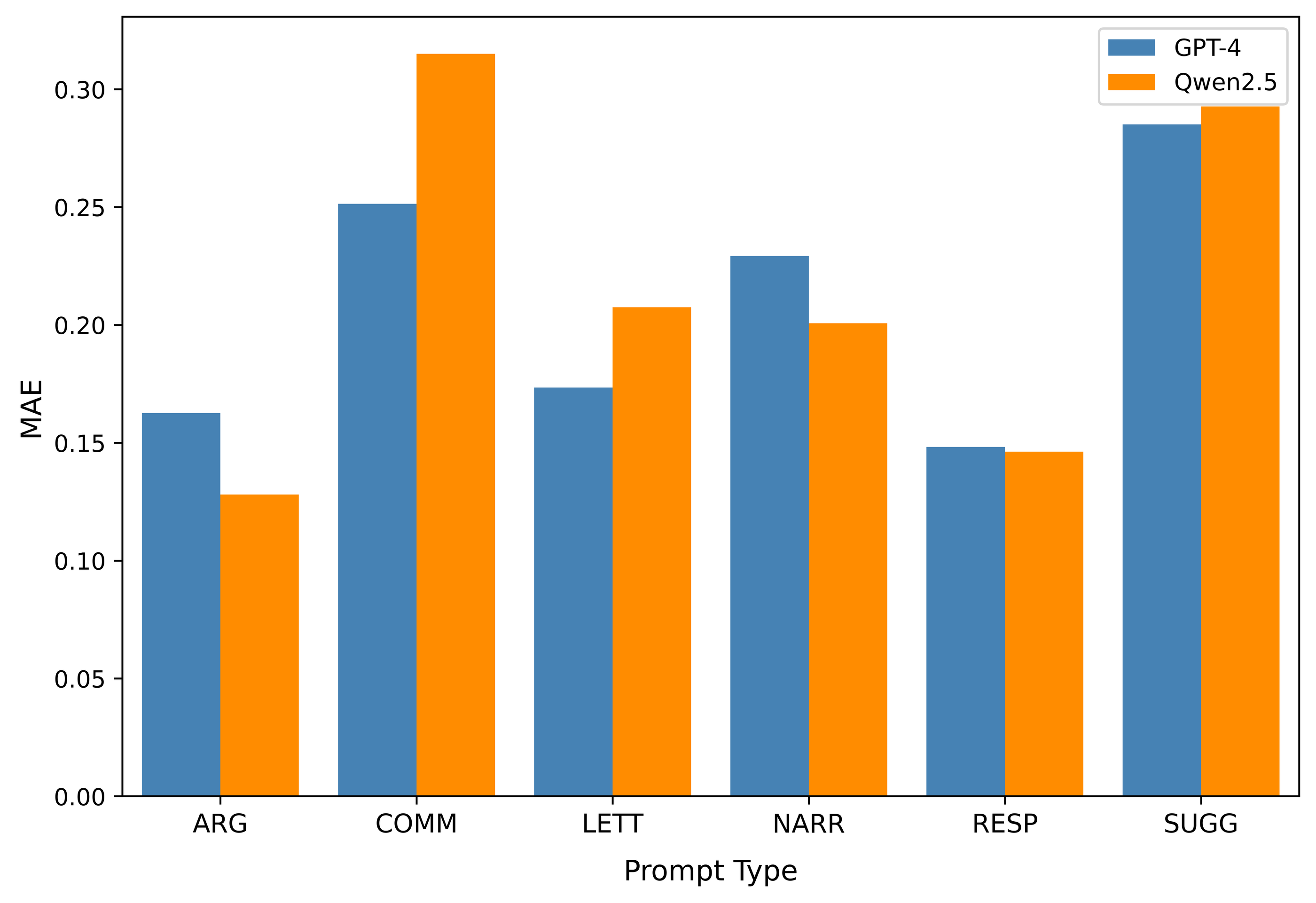}
        \label{fig:first_figuremae}
    \end{subfigure}
    
    \begin{subfigure}[t]{\columnwidth}  
        \centering
        \includegraphics[width=0.8\columnwidth]{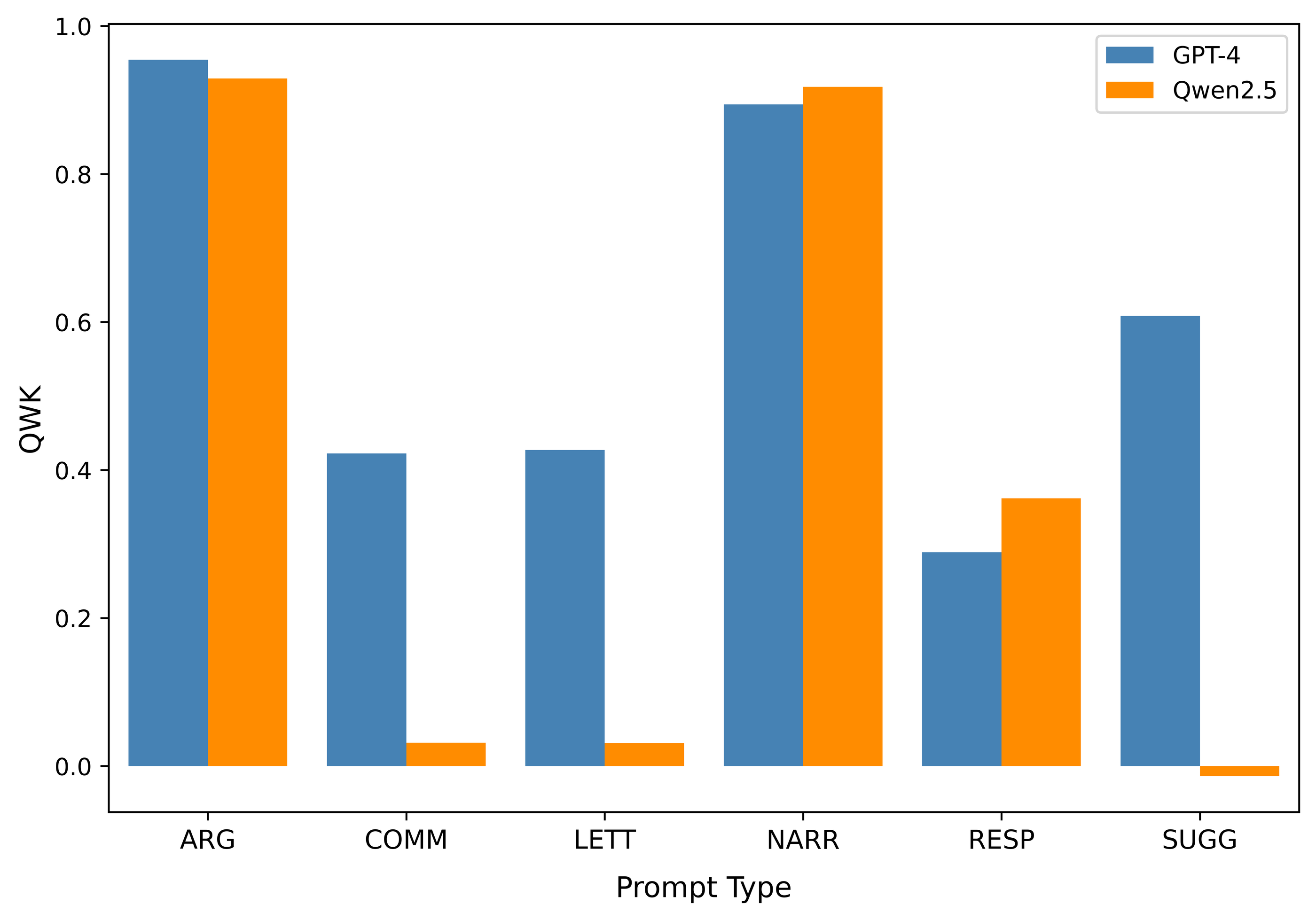}
        \label{fig:second_figureqwk}
    \end{subfigure}
    \caption{Few-shot Results Comparing GPT-4 and Qwen2.5 Across Prompt Types.}
    \label{fig:fewshotGPTQwen}
\end{figure}

In regards to the performance of GPT-4 and Qwen2.5, Figure
\ref{fig:fewshotGPTQwen} shows the MAE (top chart) and QWK (bottom chart) for
the two LLMs across each of the six prompt types. In terms of MAE, Qwen2.5's
assessment score errors are comparable to those attained by GPT-4 for most
prompt types, including response (RESP), commentary (COMM), letter (LETT),
and suggestion (SUGG) essays. GPT-4 had slightly higher error rates for
narrative (NARR), and markedly higher error when scoring 
argumentative (ARG) texts. For QWK, once again, GPT-4 and Qwen2.5 were comparable,
with GPT-4 attaining slightly better scores on letters, commentary and suggestions, while
Qwen2.5 scored higher on narratives and response. Overall, the results shed light on the
assessment performance of top closed and open LLMs for different types of
prompts and further underscore the closing performance gap between such models in the context of essay scoring.    

\begin{figure*}[t]
    \centering
    \begin{subfigure}[t]{\textwidth}
        \centering
        \includegraphics[width=\textwidth]{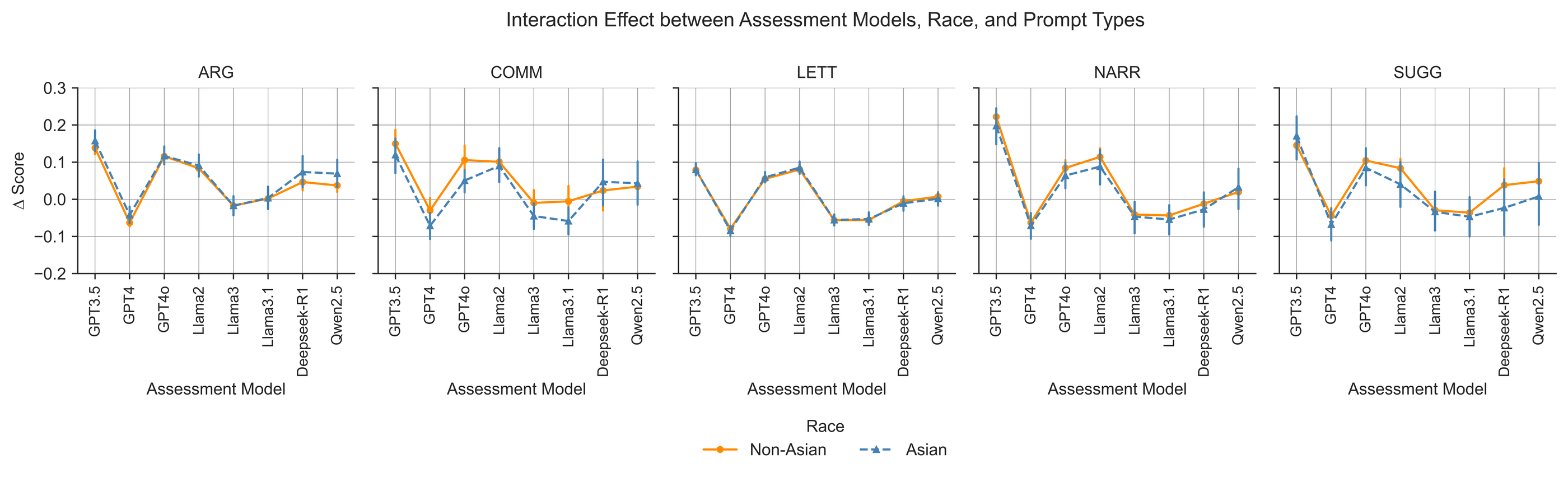}
        \label{fig:first_figurefairInt}
    \end{subfigure}
    \begin{subfigure}[t]{\textwidth}
        \centering
        \includegraphics[width=\textwidth]{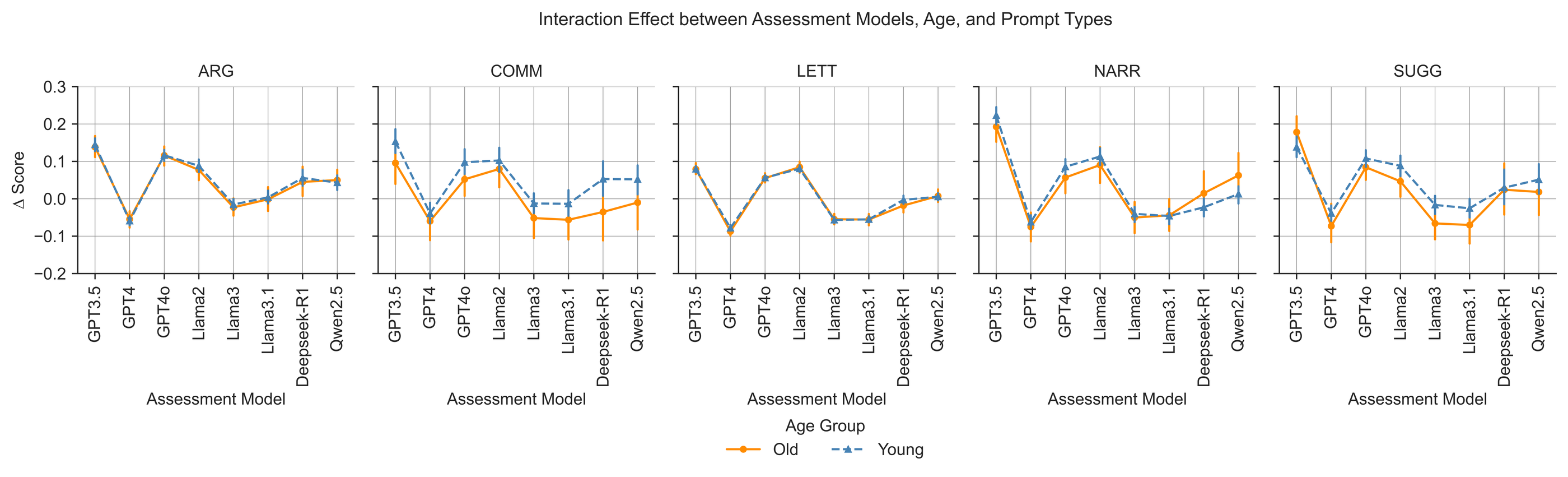}
        \label{fig:second_figurefairInt}
    \end{subfigure}
    \caption{Few-shot Results Comparing \(\Delta\) Scores (Human - LLM prediction) Across Assessment Models and Prompt Types. (left) Differences by Race, (right) Differences by Age}
    \label{fig:combinedfairInt}
\end{figure*}

\subsection{Fairness Results}

The results in Figure \ref{fig:combinedfairInt} depict the scoring error
(y-axis) for each LLM (x-axis) on a given prompt type (the five charts).
Differences between the two lines (e.g., non-Asian and Asian or older and
younger authors) indicate biases. The results reveal that all 8 LLMs excluding
OLMo 2, exhibited relatively little bias. The relative error rates for
Young/Old (bottom charts) and Asian/non-Asian (top charts) are comparable --
that is, the two sub-group lines overlay one another. This is especially true
for argument (ARG) and letter (LETT) essays. The two exceptions are
commentaries (COMM) and suggestions (SUGG), where various LLMs do exhibit biases
of up to 5\% disparate impact (i.e., differences in scoring error rates
attributable to race or age). These differences, although important to note, are
relatively mild in terms of legal, practical, and policy implications
\citep{lalor2022benchmarking,lalor2024should}. Interestingly, GPT-4 and Llama 3
exhibit similar sub-group error profiles across prompt types. In the context of
essay scoring, the results suggest that leading open LLMs may be comparable to
SOTA closed LLMs in terms of their sub-group-level bias profiles across an array
of prompt types.

\subsection{Performance of LLMs for Generation} 

\begin{figure}[tb]
        \centering
        \includegraphics[width=0.8\columnwidth]{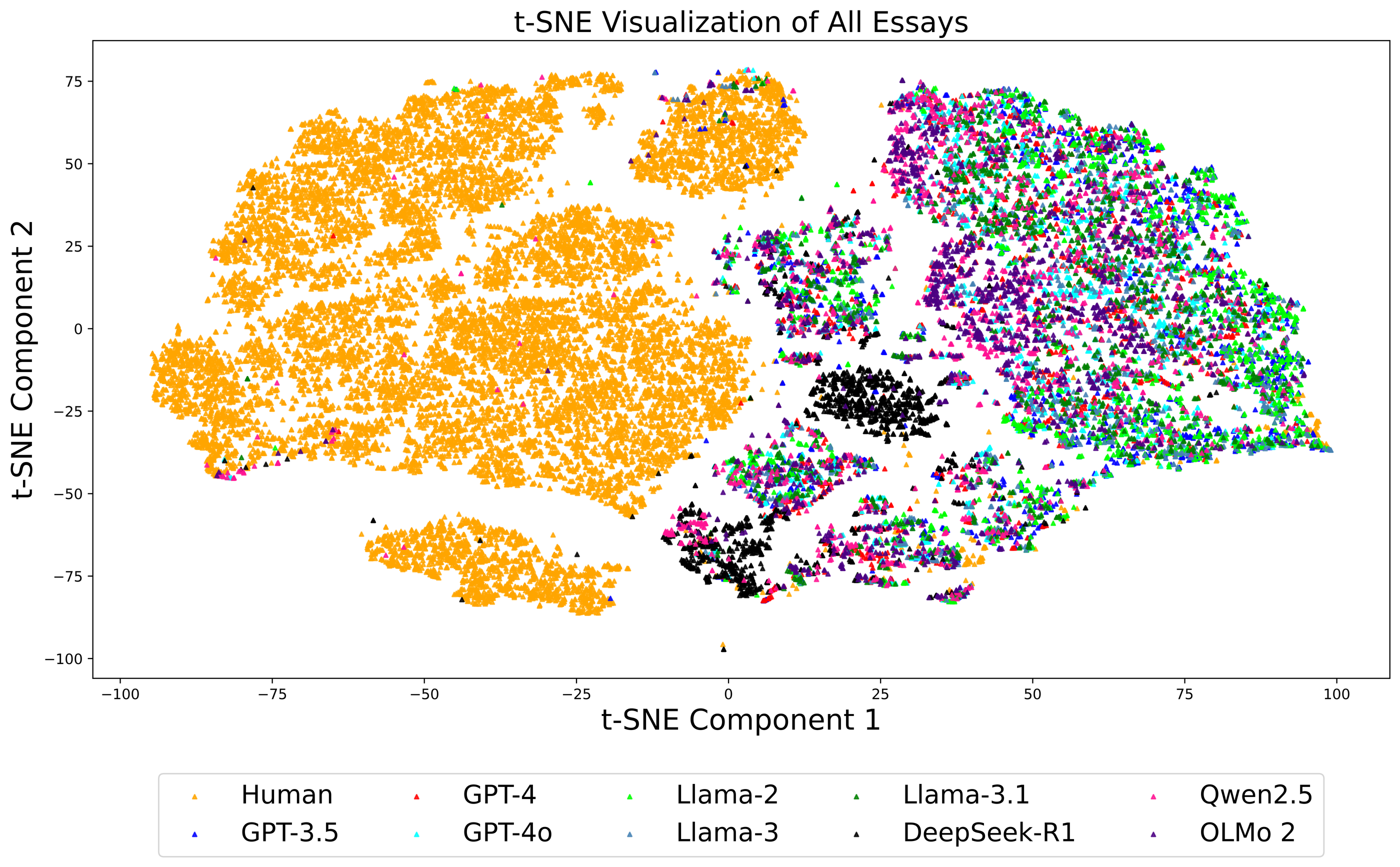}
        \caption{t-SNE plot of Human and LLM Generated Essays}
        \label{fig:tsne}
\end{figure}

Regarding RQ3, we first 
present a t-SNE (t-Distributed Stochastic Neighbor Embedding)
visualization \citep{van2008visualizing} of LLM-generated and human-written
essays based on their BERT embeddings (Figure
\ref{fig:tsne}). This visualization supports the notion
that while open and open-source LLMs like Qwen2.5 and OLMo 2 respectively, are closing the gap with
closed LLMs such as GPT-4, there remains a distinguishable difference between
machine-generated and human-written texts. The relative proximity of LLM
clusters to one another suggests that while some variability remains based on the specific model, overall these models produce essays with similar
attributes. 

\begin{figure*}[t]
    \centering
        \includegraphics[width=\textwidth]{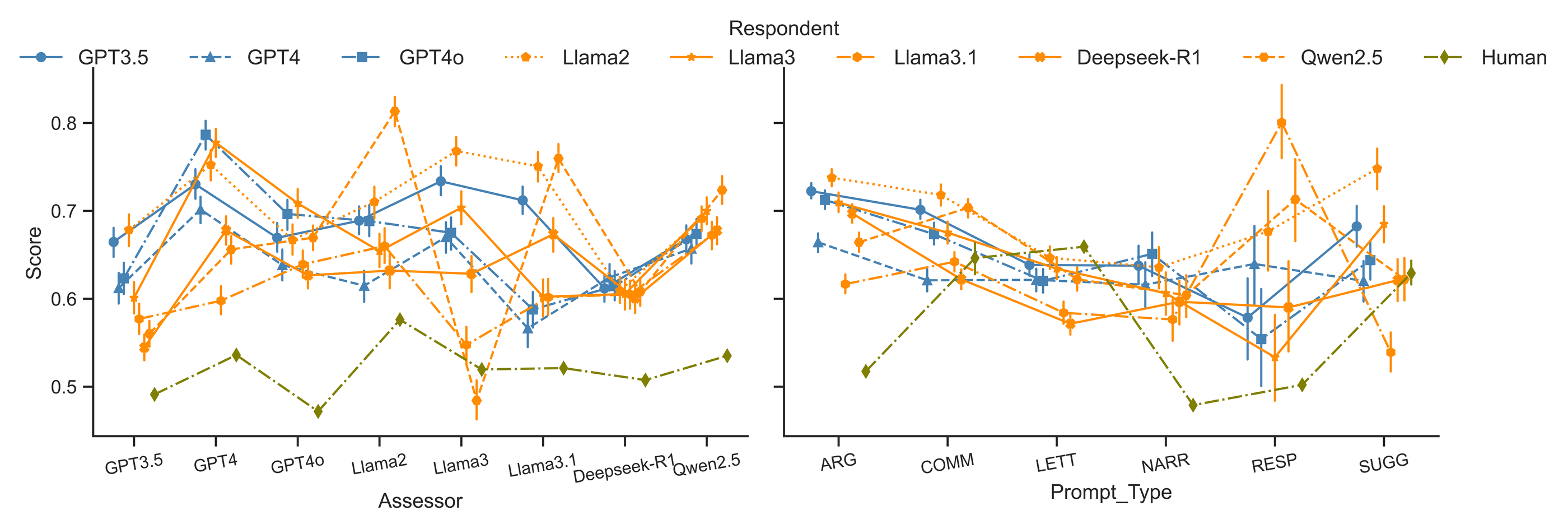}
    \caption{(left) Comparing Scores of Different LLM Assessors for LLMs/Human Generated Text, (right) Interaction Effect Between Respondent and Prompt. Blue Lines Denote Closed LLMs, Orange Denote Open LLMs}
    \label{fig:combinedresp}
\end{figure*}

\begin{table}[tbh]
\centering
\small 
\begin{tabular}{lrrrr}
  \toprule
 \textbf{Term} & \textbf{DF} & \textbf{SS} & \textbf{MS} & \textbf{F-statistic}
 \\
  \midrule
  A (Prompt Type) & 5 & 4.58e6 & 916900 & 62074.90$^{***}$ \\ 
  B (Respondent) & 9 & 2.59e6 & 288144 & 19507.59$^{***}$ \\ 
  C (Assessor) & 8 & 1.73e5 & 21674 & 1467.32$^{***}$ \\ 
  A $\times$ B & 45 & 3.68e6 & 81787 & 5537.07$^{***}$ \\ 
  A $\times$ C & 40 & 1.74e5 & 4355.04 & 294.84$^{***}$ \\ 
  B $\times$ C & 71 & 2.22e3 & 31.26 & 2.12$^{***}$ \\ 
  A $\times$ B $\times$ C & 355 & 6.54e3 & 18.42 & 1.25$^{**}$ \\ 
  \bottomrule
  \multicolumn{5}{l}{$^{***}$: $p < 0.001$}\\
\end{tabular}
\caption{Few-Shot ANOVA Results with Nine LLMs  \& Human Text.} 
\label{tab:anova-results}
\end{table}

To examine the assessment-generation interplay (RQ3), using the ANOVA model
described in Section \ref{ssec:statGen}, analysis results depicting statistical
significance for the main-effects, two-way, and three-way interactions are shown
in Table \ref{tab:anova-results}. All the factors were significant ($p< 0.05$),
suggesting that prompt-type, LLM/human respondent, and LLM assessor all
significantly impact essay assessment scores (in terms of main effects, two-way,
and three-way interactions). Figure \ref{fig:combinedresp} depicts the two-way
interactions between assessment-respondent (left chart) and
prompt-type-respondent (right chart). The assessment-respondent interactions
show that LLMs tend to rate other LLM text higher than human content (left
chart). Moreover, when looking at the assessment LLMs with the lowest prediction
error on humans, namely GPT-4, GPT-4o, Qwen2.5, and Llama 3, they tend to rate GPT-4, Qwen2.5, and
Llama 3 generated essays the highest (left chart). These results are consistent
across prompt types, with response essays (RESP) having the greatest variability
(right chart). A detailed breakdown of assessment scores is provided in Appendix A.3 (\ref{fig:assmt-gen}), illustrating these scoring trends.    

\section{Discussion and Conclusion}
This study contributes to the growing body of research exploring LLM
accessibility divides. While the emerging literature has made some strides in
evaluating the performance, bias, and costs associated with LLMs
\citep{brown2020languagemodelsfewshotlearners,openai2023gpt,touvron2023llama,bolukbasi2016man,buolamwini2018gender,raji2020closing,strubell2020energy},
our study offers an extensive, statistically robust multi-dimensional comparison
that focuses strongly on the practical and ethical implications of model choice.
The performance analyses demonstrate that while closed LLMs, particularly GPT-4,
lead in raw performance metrics, the margin is small. Open LLMs like Qwen2.5 and Llama 3
closely match GPT-4's performance. Additionally, the analysis of fairness of the
models showed that top models maintained consistent $\Delta$ scores across race
and age, indicating a low propensity for demographic bias when provided with
context (i.e., few-shot learning). 

Open LLMs such as Llama 3 offer substantial
cost savings, being up to 37 times more cost-efficient than GPT-4. This cost
advantage, combined with relatively comparable performance and fairness,
positions newer open LLMs as attractive options, particularly for those
operating with limited resources and/or in environments where greater
transparency is important.

These findings have significant implications for the NLP community. The
increasing viability of open LLMs more closely aligns with the principles of the
common-task framework. The NLP community may continue to find greater value in
adopting and contributing to open-source ecosystems, which promote innovation
while ensuring equitable access to advanced AI technologies. To conclude, this
study provides empirical evidence that challenges the dominance of closed LLMs
in recent years by demonstrating the comparative performance, fairness, and
cost-efficiency of open alternatives. Our findings underscore the democratizing
potential of SOTA open LLMs.


\section{Limitations}
Our work is not without limitations. Recent research on LLM security suggests
that open models may be more susceptible to security issues and attacks relative
to their closed counterparts. Furthermore, although open LLMs are objectively
more transparent---the inference code and tuned weights are not readily
available for closed models---the massive size of open LLMs does raise
questions about how explainable, interpretable, transparent, and scrutable
multi-billion parameter LLMs can really be \citep{bender2021dangers}.
Nevertheless, if existing in an LLM-powered world, we believe that relative to
closed models, viable open LLM alternatives capable of alleviating availability,
cost, and transparency issues are of paramount importance. Another limitation is
that, for reasons alluded to in the introduction section, we focused on the
specific context of automated essay scoring. Another benefit of AES is that by
having multiple expert raters for each data point, some of the potential
ground-truth labeling issues associated with less-structured labeling tasks
and/or single-labeler tasks were mitigated \citep{sogaard2014selection}.
Nonetheless, future work should explore other settings such as a broader array
of text-based assessment and generation tasks. 

Moreover, we chose to focus on three generations of closed and open GPT and
Llama and one generation of Qwen and DeepSeek LLMs. Other viable alternatives such as Mistral, Falcon, and so forth
could also have been included. We did so for financial/cost reasons, and to make
the ANOVA plot results more manageable and readable. Limitations
notwithstanding, our work contributes to the nascent emerging literature on LLM
accessibility divides. Our hope is that future research can build upon our work.
We intend to make all generated text, assessment data, statistical models, and
analyses scripts publicly available as a resource for future evaluation
research.

Lastly, we note that many open models (e.g., Llama
2, Llama 3) can also be downloaded and run locally. 
To ensure a fair cost
comparison, we intentionally relied on API-based services for the closed (GPT) and open
(Llama, Qwen, DeepSeek-R1) models, rather than
running them on local or cloud-based servers, as done in some prior studies
\citep{wolfe2024laboratory}. 
However, we ran the OLMo 2 open-source model locally due to their full availability. This distinction highlights key trade-offs in accessibility:
API-based models offer ease of use but involve ongoing costs, while locally run
models---whether open or open-source---require technical setup and computational
resources but eliminate API-related expenses in the long run.

\newpage
\bibliography{anthology,acl}
\bibliographystyle{acl_natbib}
\appendix
\clearpage
\section{Appendix A}
\label{sec:appendix}

\subsection{Cost Analysis}
\begin{figure}[H]
        \centering
        \includegraphics[width=\linewidth]{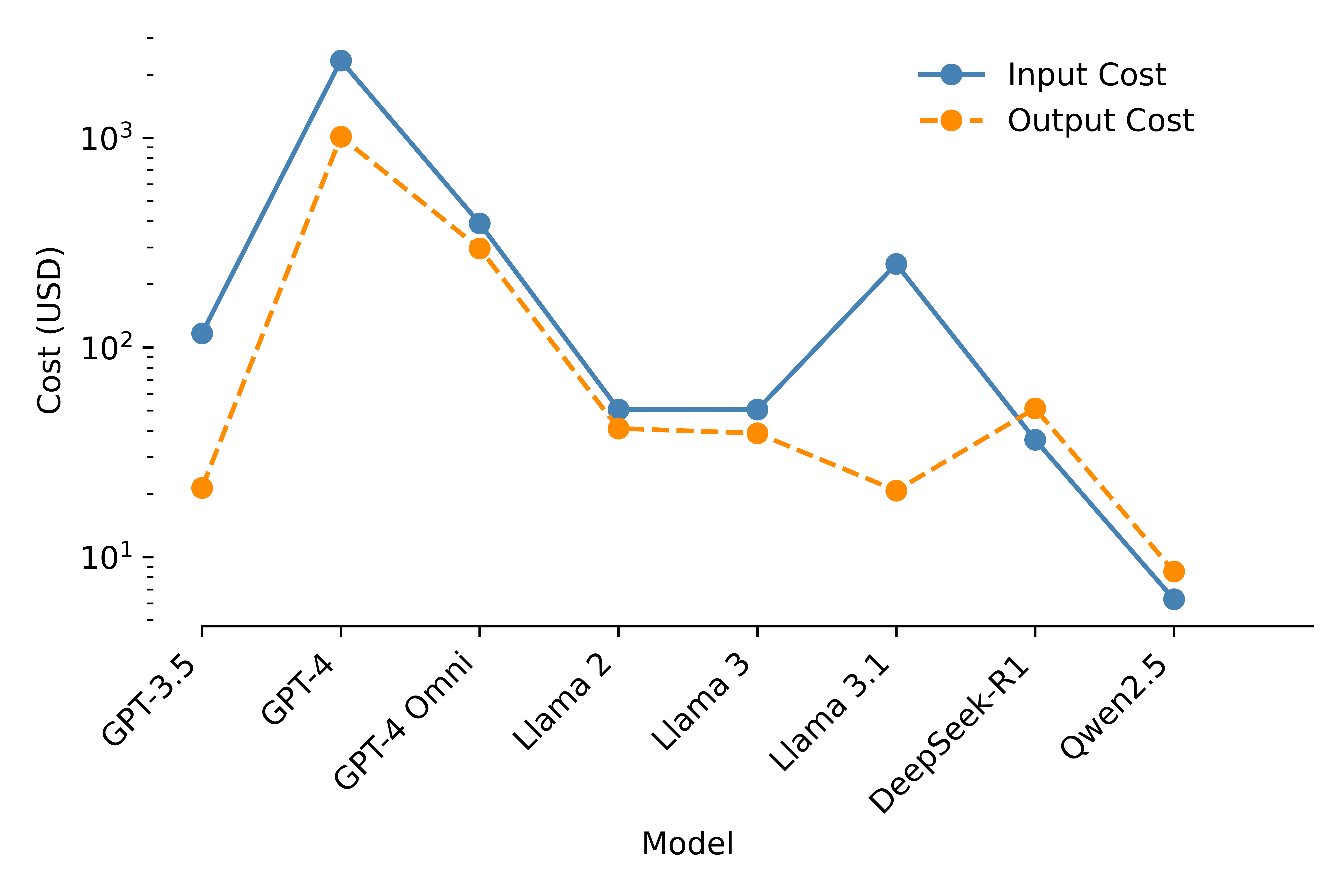}
        \caption{Input and Output Token Cost across Various LLMs. The y-axis is
        log-scaled for readability. Costs calculated as of January
        2025}
        \label{fig:costs}
\end{figure}

To compare and contrast the cost-benefit trade-offs of open vs. closed LLMs, we computed the
input and output token utilization cost of the LLMs across the assessment and
generation tasks. As noted earlier in Section 3.2, in order to allow a fair
comparison of cost, we compared the open and closed models when running both via
APIs (i.e., we used the OpenAI, Replicate, Llama, and DeepInfra APIs). Figure
\ref{fig:costs} shows the eight LLMs and the cost in thousands (in USD) associated
with input and output tokens per LLM. GPT-4 exhibits the highest input and
output costs, reflecting its substantial computational resource requirements. In
contrast, open LLMs such as Llama 3, DeepSeek-R1, and Qwen2.5 demonstrate significantly lower
costs (15-17 times lower than GPT-4), emphasizing their cost-efficiency for
comparable performance relative to closed alternatives.

\subsection{Further Few-Shot Evaluation}

\begin{figure}[H]
    \centering
    \begin{subfigure}[t]{\columnwidth}  
        \centering
        \includegraphics[width=0.8\columnwidth]{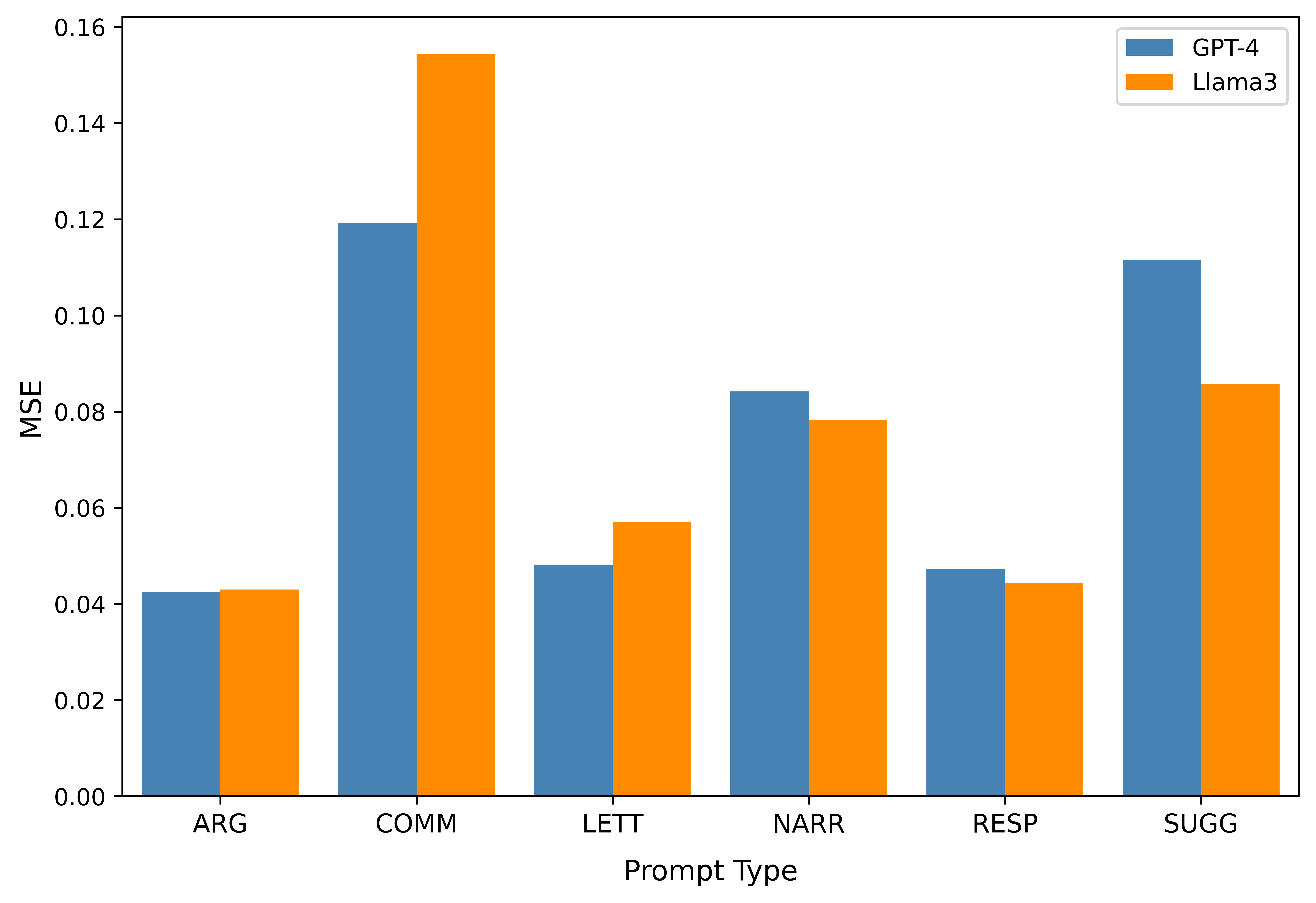}
        \label{fig:first_figureglmae}
    \end{subfigure}
    
    \begin{subfigure}[t]{\columnwidth}  
        \centering
        \includegraphics[width=0.8\columnwidth]{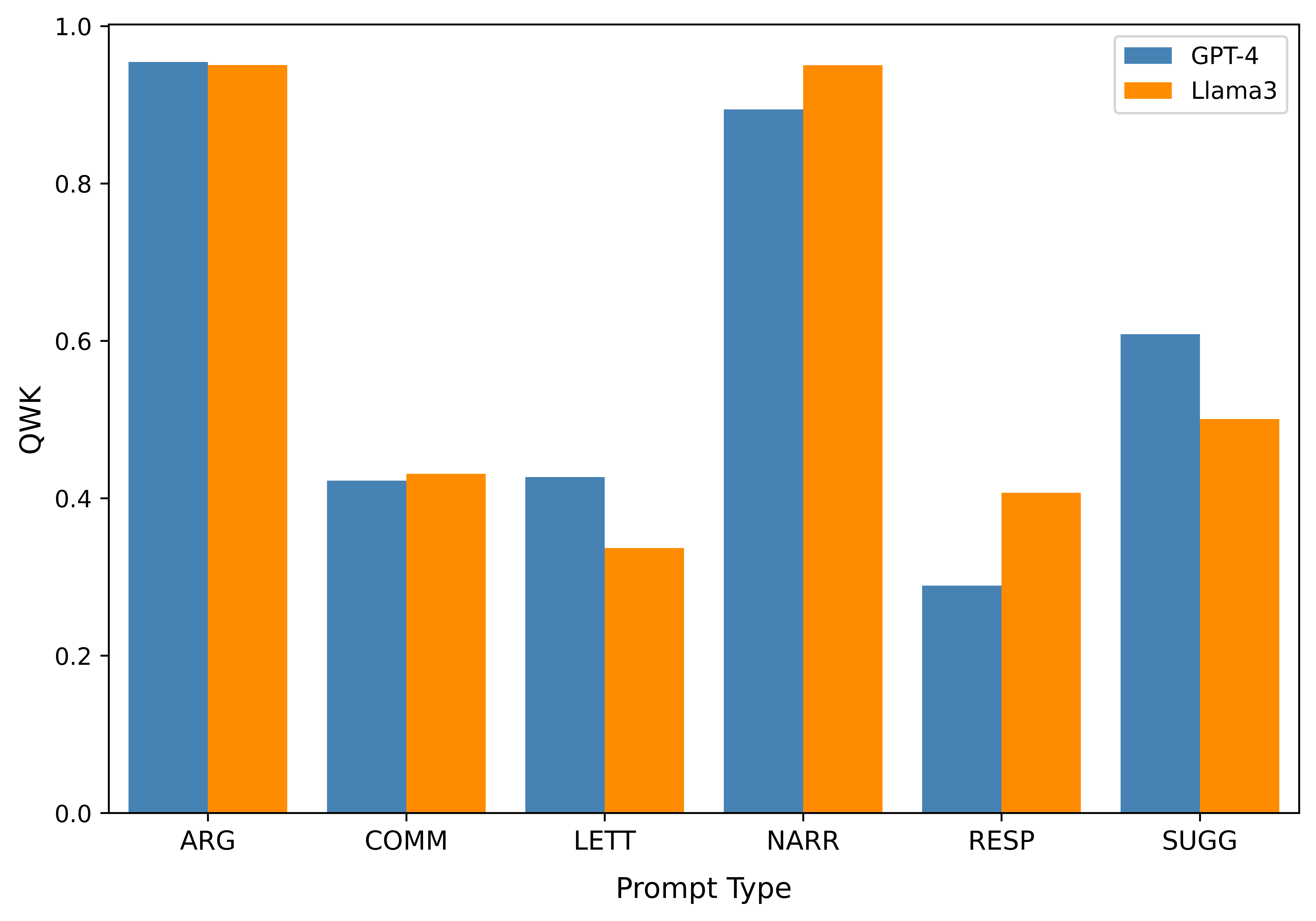}
        \label{fig:second_figureglqwk}
    \end{subfigure}
    \caption{Few-shot Results Comparing GPT-4 and Llama 3 Across Prompt Types}
    \label{fig:fewshotGPTLlama}
\end{figure}

Figure \ref{fig:fewshotGPTLlama} presents an extension of our few-shot evaluation, comparing GPT-4 and Llama 3 across different prompt types. Consistent with our findings earlier, where Qwen2.5 demonstrated strong performance relative to GPT-4, Llama 3 exhibits comparable effectiveness across multiple prompt types, further reinforcing the capability of open models. While GPT-4 maintains a slight advantage in COMM and SUGG, Llama 3 closely matches or outperforms GPT-4 in NARR, RESP, and ARG when measured by QWK. These results provide additional evidence that open LLMs are increasingly competitive with closed SOTA models.

\subsection{LLM Assessment Scores Breakdown}
\begin{figure}[H]
    \centering
    \includegraphics[width=1.0\linewidth]{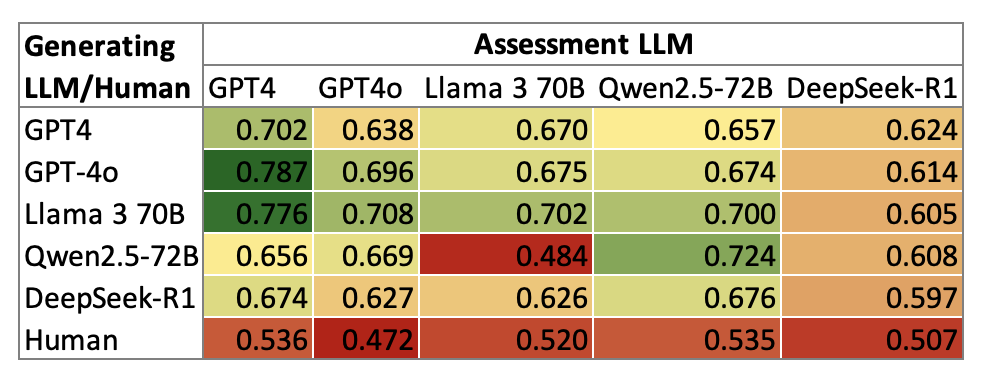}
    \caption{Average Assessment Scores of LLMs/Human-Generated Text by Different LLMs}
    \label{fig:assmt-gen}
\end{figure}

Figure \ref{fig:assmt-gen} presents average assessment scores assigned by different LLMs to essays generated by LLMs and human respondents. The red-to-green color scale highlights score variations, where green represents higher ratings and red represents lower ratings.
This visualization further supports the trends observed in Figure \ref{fig:combinedresp}, showing that LLM assessors tend to rate other LLM-generated text higher than human-written responses. 

\end{document}